%% file: main.tex
\documentclass{article}
\usepackage{graphicx}
\usepackage{amsmath}
\usepackage{booktabs}
\usepackage{multirow}
\usepackage[none]{hyphenat}
\usepackage{enumitem}
\usepackage{amssymb}
\usepackage{wrapfig}
\usepackage{xcolor} 
\usepackage[final]{corl_2026} 

\definecolor{crblue}{RGB}{0,65,180}
\newif\ifcrchanges
\crchangestrue

\title{Heterogeneous Tactile Transformer}

\author{
  Jianxin Bi$^{1\dagger}$, Qiang Wang$^{1}$, Jayaram Reddy$^{1}$, Kelvin Lin$^{1}$ \\
  \bfseries  Soibkhon Khajikhanov$^{1}$, Ruihan Gao$^{2}$, and Harold Soh$^{1,3\dagger}$ \\
  \\
  {\normalfont\small $^{1}$National University of Singapore, $^{2}$Carnegie Mellon University, $^{3}$Smart Systems Institute, NUS} \\
  {\normalfont $^{\dagger}$\ Corresponding authors}
}

\hypersetup{pdftitle={Heterogeneous Tactile Transformer},pdfauthor={Jianxin Bi, Qiang Wang, Jayaram Reddy, Kelvin Lin, Soibkhon Khajikhanov, Ruihan Gao, Harold Soh}}
\begin{document}
\maketitle


\begin{abstract}
Tactile sensors are inherently heterogeneous: a model trained on one sensor cannot be directly used on another, which limits learning contact-rich manipulation policies from diverse tactile data at scale. To bridge this gap, we propose the Heterogeneous Tactile Transformer (HTT), a framework that learns shared tactile representations across heterogeneous sensors. HTT consists of sensor-specific encoders and a shared transformer trunk, and is pretrained with per-modality masked reconstruction together with cross-modal alignment between paired sensors. Pretraining uses our novel  Heterogeneous Paired Tactile (HPT) dataset, containing 1.6M synchronized paired frames across four vision- and array-based tactile sensors. Across distinct tactile perception and real-world manipulation tasks, HTT is shown to learn transferable representations that adapt to new tasks and previously unseen sensors. Dataset, code, and model checkpoints are released at \href{https://jxbi1010.github.io/htt-gh-page/}{this link}.
\end{abstract}

\keywords{Tactile Representation, Contact-Rich Manipulation} 


\input{sections/1_introduction}
\input{sections/2_related_work}
\input{sections/3_dataset}
\input{sections/4_methodology}
\input{sections/5_experiments}
\input{sections/6_conclusion}


\clearpage
\acknowledgments{This research was funded by the NUS Artificial Intelligence Institute (NAII) seed grant number H1FY2025 and a gift grant from Google.}


\bibliography{references}

\clearpage
\input{sections/10_appendix}

\end{document}

%% file: sections/1_introduction.tex
\section{Introduction}

Tactile sensing is critical for robot manipulation, especially in contact-rich tasks such as grasping fragile objects, precision assembly, and material identification. Yet tactile learning remains challenging to scale. One critical bottleneck is that, unlike cameras, tactile sensors are \emph{heterogeneous} and differ in how they measure contact and in the data they produce. Optical tactile sensors such as GelSight-style devices \cite{Yuan2017GelSightHR,Lambeta_2020,lin20239dtact} capture detailed spatial and geometric information from elastomer deformation. In contrast, array-based tactile sensors \cite{xela,bhirangi2021reskin,tacniq,huang20243dvitac} provide high-rate temporal signals and more direct force or pressure measurements.

These sensing modalities are complementary, but they are not easy to combine. Optical sensors typically provide rich spatial observations but are often limited by camera frame rates. Array-based sensors provide faster force-sensitive measurements, but with lower spatial resolution. More importantly, their raw outputs have very different structures: images in one case and time series over sensing elements in the other. This mismatch makes it difficult to train reusable tactile representations or sensor-agnostic manipulation policies.

In this work, we seek a \emph{tactile backbone} for \emph{heterogeneous} sensors. Recent tactile representation learning has made strong progress, but it has largely focused on optical tactile sensors \cite{higuera2024sparsh,zhao2024transferable,gupta2025sensorinvariant,fenganytouch,yang2024binding}. Self-supervised methods based on Masked Autoencoders (MAE) \cite{mae} can learn powerful features from GelSight-style images \cite{zhao2024transferable, gupta2025sensorinvariant}. However, models trained only on optical tactile data inherit the limits of that sensor: they cannot directly ingest array-based tactile signals, and they may be less suitable for tasks that depend on high-rate force or slip cues. This motivates a heterogeneous tactile backbone that aligns sensor representations in a shared latent space, while preserving the structure and strengths of each sensor class.

To this end, we introduce the \textbf{Heterogeneous Tactile Transformer (HTT)}, a self-supervised framework trained on our curated \textbf{Heterogeneous Paired Tactile (HPT)} dataset. HPT contains $1.6$M paired tactile frames from optical sensors (GelSight Mini, 9DTact) and array-based sensors (Xela, Tac-02), collected using a Universal Manipulation Interface (UMI) setup \cite{chi2024universal,YuS-RSS-25}. The dataset covers synchronized sensor-object interactions across \textit{press}, \textit{twist}, and \textit{slide} motions. HTT uses sensor-specific encoders to process each sensor and a shared transformer trunk to form a common latent space. During pretraining, it combines masked reconstruction with bidirectional cross-sensor prediction. This allows the model to preserve sensor-specific structure while aligning representations across sensors.

We evaluate HTT on three tactile perception tasks: \textit{object classification}, \textit{force estimation}, and \textit{slip detection}. We further evaluate it on simulated and real-world robot manipulation tasks. Across these settings, HTT learns transferable representations for heterogeneous tactile sensors and improves downstream policy learning with novel tactile inputs. In particular, HTT-derived embeddings benefit contact-rich tasks such as \textit{toy screw} and \textit{grasp tofu}. These results suggest that paired heterogeneous pretraining is a practical path toward more general tactile backbones.

In summary, our work studies how to learn unified tactile representations across heterogeneous sensors, with the goal of improving transfer to downstream tasks and novel sensor types. Our core contributions are as follows:
\begin{itemize}[noitemsep,topsep=1mm]
    \item \textbf{Heterogeneous Paired Tactile Dataset (HPT)}, a large-scale tactile dataset containing $1.6$M paired frames from optical and array-based tactile sensors. HPT provides synchronized cross-sensor data for studying heterogeneous tactile representation learning.
    \item \textbf{Heterogeneous Tactile Transformer (HTT)}, a self-supervised framework that combines MAE-style masked reconstruction with bidirectional cross-sensor prediction to learn shared representations across heterogeneous tactile sensors.
    \item \textbf{Comprehensive Evaluation} across tactile perception tasks, simulated manipulation tasks, and real-world robot experiments. The results show that HTT learns transferable tactile representations and can serve as a general backbone for heterogeneous tactile sensing.
\end{itemize}

\begin{figure}
    \centering
    \includegraphics[width=0.98\textwidth]{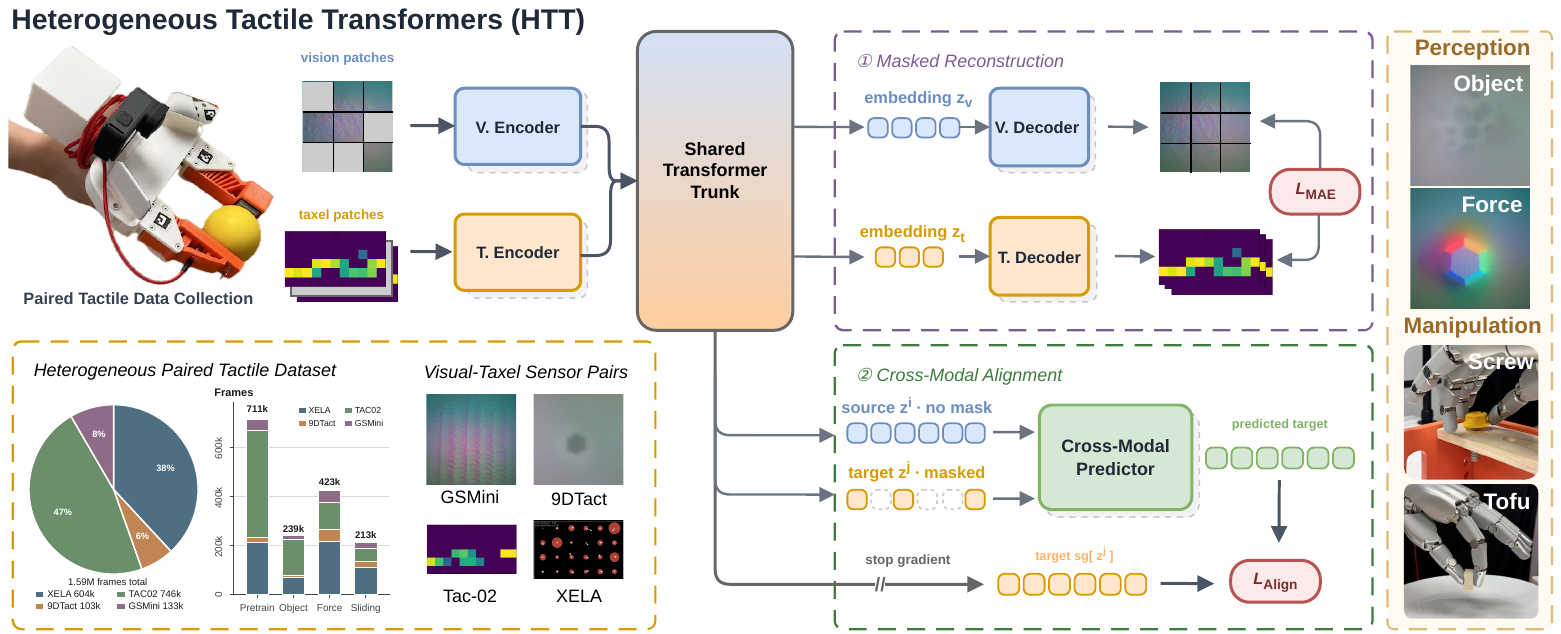} 
    \caption{\small\textbf{Heterogeneous Tactile Transformer (HTT).} HTT is pretrained on the \textit{Heterogeneous Paired Tactile (HPT) dataset}, which consists of $1.6$M synchronized paired frames across four distinct sensors with a UMI device. HTT adopts sensor-specific encoders and a shared transformer trunk as core modules. Data from each sensor is patchified, fed into a sensor-specific encoder, and forwarded to the shared transformer trunk. During pretraining, decoders are used to reconstruct each sensors' input data and cross-sensor predictors are used to predict masked target sensor embedding  --- aligning the shared latent space across heterogeneous sensors. The pretrained HTT model can be applied to distinct perception tasks and boost manipulation policy learning on unseen sensors.}
    \label{fig:HTT}
    \vspace{-12pt}
\end{figure}

%% file: sections/2_related_work.tex
\section{Related Work} \label{Sec: related works}
This section gives a brief overview of tactile sensing, datasets, and representation learning for heterogeneous sensors; broader reviews of tactile sensing and tactile perception for robotics can be found in survey articles (e.g., \cite{li2020review,luo2017robotic,li2024comprehensive}).

\textbf{Tactile Sensors.}
Tactile sensors differ substantially in their transduction mechanisms and output formats. In this work, we focus on two broad families that are common in robot learning: optical tactile sensors and array-based tactile sensors. Optical tactile sensors, such as GelSight~\cite{Yuan2017GelSightHR, gelsightmini}, DIGIT~\cite{9018215, lambeta2024digitizingtouchartificialmultimodal}, 9DTact~\cite{lin20239dtact}, ThinTac~\cite{xu2025thintact}, and PolyTouch~\cite{zhao2025polytouchrobustmultimodaltactile}, infer contact properties from images of a deformable contact surface. These sensors provide rich spatial information for texture, shape, and contact geometry, but are often limited by camera frame rates and sensor-to-sensor variation. Array-based tactile sensors, such as Xela uSkin~\cite{xela}, PapillArray~\cite{contactile}, and TAC-02~\cite{tacniq}, measure contact through distributed sensing elements that convert physical stimuli into electrical signals. They provide high-rate force- or pressure-sensitive measurements but have lower spatial resolution than optical tactile sensors. These complementary properties motivate learning representations from both sensor families.

\textbf{Tactile Datasets for Robot Learning.}
Large-scale tactile representation learning depends on datasets that cover diverse sensors, objects, and interactions. 
Most public tactile datasets used for robot learning remain centered on optical tactile sensors~\cite{yang2022touch, zhao2024transferable,higuera2024sparsh,fenganytouch,suresh2022midastouch,yu2024octopi, fenganytouch2}; some provide paired observations across different optical sensors~\cite{fenganytouch}, while others leverage physics-based simulation~\cite{gupta2025sensorinvariant, si2024difftactile} or implicit neural representations~\cite{gao2022objectfolder} to increase scale and diversity.
However, they largely remain within the optical tactile setting. To our knowledge, there is still no large-scale dataset of temporally paired optical and array-based tactile observations at this scale. We address this gap by introducing a low-cost paired data collection method and a large-scale dataset for heterogeneous tactile representation learning.

\textbf{Heterogeneous Tactile Representation Learning.}
Most existing tactile representation learning methods focus on optical tactile sensors~\cite{higuera2024sparsh,zhao2024transferable,gupta2025sensorinvariant,xu2025unit,fenganytouch2}, often using self-supervised reconstruction \cite{higuera2024sparsh,zhao2024transferable} or contrastive learning \cite{gupta2025sensorinvariant}. Other works align tactile data with non-tactile modalities such as vision \cite{kerr2022self,fenganytouch,fu2024a,jones2025sightfinetuninggeneralistrobot}, language \cite{yu2024octopi,fenganytouch,fu2024a}, and audio \cite{higuera2025tactile}. Fewer works study representation learning across heterogeneous tactile sensors. UniTac \cite{hou2025unitac} learns a shared representation across non-vision tactile sensors, prior work has fused optical and array-based tactile data for supervised, task-specific learning \cite{gao2020supervised,Zandonati2023InvestigatingVF}. Touch-to-touch translation \cite{grella2025touch} learns a one-way mapping between tactile sensing technologies, but does not target a general reusable representation. UniForce \cite{chen2026uniforceunifiedlatentforce} also collects synchronized heterogeneous tactile data using UMI, but focuses on latent force prediction rather than general tactile pretraining. In contrast, HTT learns a shared self-supervised representation across optical and array-based tactile sensors via paired-data cross-modal alignment, and evaluates transfer across perception and manipulation tasks.

%% file: sections/3_dataset.tex
\section{Heterogeneous Paired Tactile Dataset } \label{Sec: dataset}

To address the lack of paired optical-based and taxel-based tactile data, we design a custom data acquisition system based on a Universal Manipulation Interface (UMI) ~\cite{chi2024universal}. 
This configuration facilitates the collection of synchronous, heterogeneous tactile data during real-world sensor-object interactions. Our system utilizes 3D-printed modular shells to securely mount two different tactile sensors on opposing sides of the gripper fingers. This opposing layout allows for the simultaneous measurement of contact properties during a single interaction episode. 
We define two primary sensor configurations for data collection: \textit{Pair A} (\textit{Xela} $\leftrightarrow$ \textit{9DTact}) and \textit{Pair B} (\textit{TAC-02} $\leftrightarrow$ \textit{GS Mini}). 

In our data collection procedure, pairing is synchronized at the grasp level, rather than calibrated pointwise across contact surfaces. Objects are selected to have near-symmetric geometry under the grasp pose, so opposing contacts share coarse texture, local shape, and force attributes. This supports collection without per-contact calibration fixtures, making such paired data  easier to collect at scale.

\begin{figure}[t]
    \centering
    \includegraphics[width=1.0\textwidth]{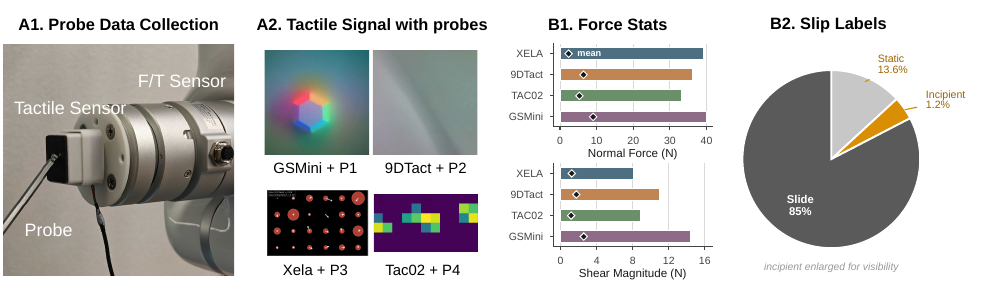}
    \vspace{-20pt}
    \caption{\textbf{Force/slip data collection and dataset statistics.} \textbf{A1.} A tactile sensor and a $6$-D F/T sensor are mounted on a robot arm; a probe rig contacts the tactile sensor while synchronized tactile frames and ground-truth force are recorded for the force-estimation and slip-detection splits. \textbf{A2.} Example tactile frames collected with the four probe geometries. \textbf{B1.} The force range spans up to $40$\,N normal and $14$\,N shear. \textbf{B2.} Slip labels are heavily imbalanced ($13.6\%$ static, $1.2\%$ incipient, $85.2\%$ slide), making the rare static and incipient classes challenging to detect.}
    \label{Fig: dataset}
    \vspace{-12pt}
\end{figure}

\subsection{Dataset Components and Structure} We collect a large set of interaction data on daily-life objects, totaling $1.6$M tactile frames across the four sensors, with four distinct components designed to support both self-supervised pre-training and downstream perception evaluation. Note that there is \emph{no overlap} between unlabeled pretraining data and evaluation data.

\textbf{Paired Unlabeled Interaction Data.}
This dataset forms the core of heterogeneous tactile pretraining. It is collected using UMI via unscripted interactions on a diverse set of daily-life indoor objects, where each interaction yields synchronized paired streams from both sensors. The pretraining set contains 30 objects varying in hardness, roughness, bumpiness, and shape.

\textbf{Object Classification.}
We use UMI to interact with 20 distinct objects spanning diverse geometries, textures, and materials. In each episode, we choose an action from \textit{\{press, twist, slide\}} to interact with the objects to elicit varied contact patterns. The full object list can be found in Appendix~\ref{App: dataset}.

\textbf{Force Estimation.}
We collect tactile-force data with a probing rig in which an external F/T sensor attached to the tactile sensor provides synchronized $6$-D force labels (Figure~\ref{Fig: dataset}.A1). The rig uses four distinct probe geometries (Figure~\ref{Fig: dataset}.A2) to elicit varied contact patterns.

\textbf{Slip Detection.} Following the same setup as force data collection, we use probes to slide on tactile sensors and derive per-frame slip labels post-hoc from the $6$-D force by computing a friction-coefficient time series and apply a two-sided Page CUSUM change-point detector~\cite{page1954continuous}, yielding three classes: \textit{static}, \textit{incipient}, and \textit{gross slip} with class distribution shown in (Figure~\ref{Fig: dataset}.B2).
Exact sliding labeling formulation are deferred to Appendix~\ref{App: dataset}.

%% file: sections/4_methodology.tex
\section{Heterogeneous Tactile Transformer } \label{Sec: method}
The Heterogeneous Tactile Transformer (HTT) framework (Fig. \ref{fig:HTT}) is designed to learn a shared, transferable representation from heterogeneous tactile sensors via self-supervised pretraining. The core objective is twofold: (i) to enable each sensor to extract robust features from its raw signal via masked reconstruction, and (ii) to leverage paired cross-sensor data to align disparate modalities within a shared latent space via cross-modal prediction.

\subsection{HTT Model Architecture}
HTT adopts a Masked Autoencoder (MAE) design~\cite{mae} comprising sensor-specific encoders and decoders, a shared transformer trunk, and cross-modal predictors. To ensure structural compatibility across the network, embeddings from all sensors are projected to a uniform embedding space.

\textbf{Input Patching.}
Each training instance comprises synchronized multi-sensor data captured within a fixed $\tau = 0.2\,\text{s}$ temporal interaction window. To accommodate the heterogeneous input shapes, we apply sensor-specific tokenization pipelines. For optical-based sensors, individual frames are resized to $224 \times 224$ and tokenized into non-overlapping spatial patches following standard Vision Transformer (ViT)~\cite{dosovitskiy2021vit}. For array-based sensors, the high-frequency multi-dimensional time series is tokenized into non-overlapping temporal patches. For all modalities, \emph{we subtract raw tactile signal with a non-contact reference frame before processing by model.} Detailed patch dimensions, sequence lengths, and per-sensor frame counts within the window are provided in Appendix~\ref{App: method}.

\textbf{Architecture.} Let $\mathcal{I}$ denote the set of available tactile sensors, where $i \in \mathcal{I}$. The architecture consists of four primary components:

\begin{itemize}[noitemsep, topsep=2pt, parsep=0pt]
    \item \textbf{Encoders ($\mathcal{E}_i$):} Sensor-specific encoders that encode tactile tokens to embeddings (ViT for optical, self-attention transformer for array-based tactile signal).
    \item \textbf{Shared Trunk ($\mathcal{T}$):} A unified transformer trunk that processes token embeddings in a shared latent space.
    \item \textbf{Decoders ($\mathcal{D}_i$):} Sensor-specific decoders that handle masked reconstruction.
    \item \textbf{Predictors ($\mathcal{P}_{ij}$):} Cross-attention transformer that predict masked target embeddings from source embeddings and visible target embeddings.
\end{itemize}

\subsection{HTT Pretraining}
\textbf{MAE Reconstruction.} 
To encourage each sensor to extract structural and physical features from its raw signal, we apply an MAE-style masked reconstruction objective on every modality. The sensor-specific encoder $\mathcal{E}_i$ and shared trunk $\mathcal{T}$ process \emph{only} the visible tokens $\mathbf{x}^i_v$, and the sensor-specific decoder $\mathcal{D}_i$ reconstructs the masked tokens $\mathbf{x}^i_m$. The loss is averaged over modalities:

\begin{equation}
\mathcal{L}_{\text{MAE}} = \mathbb{E}_{i \sim \mathcal{I}} \left[ \left\| \mathcal{D}_i\bigl(\mathcal{T}(\mathcal{E}_i(\mathbf{x}^i_v))\bigr) - \mathrm{norm}(\mathbf{x}^i_m) \right\|_2^2\right],
\end{equation}

where $\mathrm{norm}(\cdot)$ denotes the per-patch normalization (zero-mean, unit-variance) applied to the target as in standard MAE training~\cite{mae}. This reconstruction constraint forces the encoder-trunk system to retain high-fidelity localized details necessary to restore missing spatial or temporal signals.

\textbf{Cross-Modal Alignment.}
For each ordered sensor pair $(i, j)$, the model is trained to predict the masked target embeddings from the full source embeddings and the visible target embeddings. Let $\mathbf{z}^i = \mathcal{T}(\mathcal{E}_i(\mathbf{x}^i))$ and $\mathbf{z}^j = \mathcal{T}(\mathcal{E}_j(\mathbf{x}^j))$ be the source and target embeddings, with the target partitioned into visible and masked components $\mathbf{z}^j_v$ and $\mathbf{z}^j_m$. The cross-attention predictor $\mathcal{P}_{ij}$ takes the source $\mathbf{z}^i$ together with the visible target $\mathbf{z}^j_v$, and predicts the masked target $\mathbf{z}^j_m$. To prevent representation collapse, a stop-gradient $\mathrm{sg}[\cdot]$ is applied to the regression target. The alignment loss is averaged over the set $\mathcal{S}$ of all ordered sensor pairs in the dataset, the total alignment loss is: 

\begin{equation}
\mathcal{L}_{\text{Align}} = \mathbb{E}_{(i, j) \sim \mathcal{S}} \bigl\| \mathcal{P}_{ij}(\mathbf{z}^i, \mathbf{z}^j_v) - \mathrm{sg}[\mathbf{z}^j_m] \bigr\|_2^2 .
\end{equation}

\textbf{Joint Pretraining.}
The total pretraining loss combines per-modality reconstruction and cross-modal alignment:

\begin{equation}
\mathcal{L}_{\text{HTT}} = \mathcal{L}_{\text{MAE}} + \alpha_t \cdot \mathcal{L}_{\text{Align}} ,
\end{equation}

where $\alpha_t$ is a time-dependent coefficient: $\alpha_t = 0$ during an initial warmup period so the encoders and trunk first develop sensor-specific features via MAE alone, after which $\alpha_t$ is ramped to its maximum value $\alpha_{\max} = 0.1$. To further protect those features, we block alignment gradients at the encoder outputs, so that $\mathcal{L}_{\text{Align}}$ updates only the predictors $\mathcal{P}_{ij}$ and the shared trunk $\mathcal{T}$, while the encoders $\mathcal{E}_i$ are updated solely by $\mathcal{L}_{\text{MAE}}$. 
After pretraining, the decoders $\mathcal{D}_i$ and predictors $\mathcal{P}_{ij}$ are discarded; only the sensor encoder $\mathcal{E}_i$ and shared trunk $\mathcal{T}$ are used for downstream tasks. Detailed pretraining hyperparameters are in Appendix~\ref{App: method}.

%% file: sections/5_experiments.tex
\section{Experiments and Analysis} \label{Sec: experiments}
We evaluate HTT on tactile perception tasks and robot manipulation tasks to answer the following questions:
\begin{itemize}[noitemsep, topsep=2pt, parsep=0pt]
    \item \textbf{Q1.} Does HTT pretraining yield useful representation from heterogeneous sensors?
    \item \textbf{Q2.} Does HTT representation transfer to distinct tasks? 
    \item \textbf{Q3.} How does cross-sensor alignment benefit representation learning?
    \item \textbf{Q4.} Does HTT boost contact-rich policy learning? 
    \item \textbf{Q5.} Does HTT adapt to new sensors that were not seen in pretraining?
\end{itemize}

\subsection{Experimental Design} \label{Sec: setup}

\textbf{Tasks and Data.}
We organize the evaluation into three groups of tasks aligned with Q1--Q5, with the grouping partly dictated by \emph{how the data was collected}.
\textit{Object classification} uses the same UMI paired-collection setup as the pretraining data (Section~\ref{Sec: dataset}), differing only in the interacted objects and the presence of object labels. This matched distribution makes it the cleanest probe of whether HTT pretraining yields useful features across all four heterogeneous sensors (Q1, Q3).
\textit{Force estimation} and \textit{slip detection} use the controlled $4$-probe rig, instead of the paired UMI interaction. This distribution shift lets us test whether the learned representation transfers to physics-grounded tasks beyond the pretraining setup (Q2, Q3).
The real-world robot experiments and the ManiFeel simulated tasks evaluate whether HTT embeddings boost downstream policy learning on contact-rich manipulation tasks (Q4). Critically, robot experiments use tactile sensors that are \emph{not} seen in pretraining, which tests whether HTT generalizes beyond its four pretraining sensors and can serve as a reusable backbone for new tactile sensors (Q5).

\textbf{Compared Methods.}
We compare the following methods.
\textbf{Scratch} trains sensor- and task-specific transformers with MLP heads, without pretraining.
\textbf{T3}~\cite{zhao2024transferable} and \textbf{SITR}~\cite{gupta2025sensorinvariant} are pretrained tactile representation models that target optical-based sensors only: T3 is pretrained on a large real-world tactile dataset, and SITR mainly on a synthesized optical-based dataset.
\textbf{AnyTouch}~\cite{fenganytouch} learns static--dynamic representations from optical tactile images and videos. We evaluate above methods on 9DTact and GSMini as they support visual tactile data in nature.
\textbf{SPH} (supervised heterogeneous encoder) trains HTT's encoders and shared trunk on downstream supervision only, without pretraining. Unlike sensor-specific Scratch, it controls for the heterogeneous architecture.
\textbf{MAE (ours)} uses the same architecture as HTT but is only pretrained with MAE reconstruction loss.
\textbf{HTT (ours)} additionally adds the cross-modal alignment loss on paired tactile data.
For each task we finetune the pretrained encoder $\mathcal{E}_i$ and shared trunk $\mathcal{T}$ jointly with a task-specific head or policy. Full details are in Appendix~\ref{App: baselines}.
For T3, SITR and AnyTouch, we use official checkpoints and original preprocessing approaches respectively; their image encoders cannot directly ingest array-based tactile time series, so there results is 'n/a' in table for array-based sensors. MAE and HTT share pretraining data, isolating the alignment objective.

\subsection{Object Classification Across Heterogeneous Sensors}

\begin{table}[htbp]
\centering
\setlength{\tabcolsep}{6pt}
\small
\begin{tabular}{l ccccc}
\toprule
\textbf{Method} & \textbf{Xela} & \textbf{TAC-02} & \textbf{9DTact} & \textbf{GSMini} & \textbf{Overall} \\
\midrule
Scratch     & 48.90 $\pm$ 1.45             & 22.49 $\pm$ 0.78             & 65.63 $\pm$ 1.73             & 53.13 $\pm$ 2.60             & 47.54 $\pm$ 1.64             \\
T3          & n/a                          & n/a                          & 51.44 $\pm$ 26.25            & 59.26 $\pm$ 5.25             & 55.35 $\pm$ 15.75            \\
SITR        & n/a                          & n/a                          & 81.34 $\pm$ 4.41             & 74.31 $\pm$ 7.14             & 77.83 $\pm$ 5.78             \\
AnyTouch & n/a & n/a & 71.07 $\pm$ 0.51 & 76.36 $\pm$ 5.84 & 73.72 $\pm$ 3.18 \\
SPH & 49.72 $\pm$ 0.99 & 25.29 $\pm$ 1.86 & 82.23 $\pm$ 4.05 & 75.86 $\pm$ 3.91 & 58.28 $\pm$ 2.70 \\
MAE (ours)  & \textbf{56.68 $\pm$ 0.99}    & \underline{26.16 $\pm$ 0.43} & \underline{90.08 $\pm$ 0.54} & \underline{88.59 $\pm$ 0.71} & \underline{65.38 $\pm$ 0.67} \\
HTT (ours)  & \underline{52.41 $\pm$ 0.84} & \textbf{26.20 $\pm$ 1.60}    & \textbf{94.84 $\pm$ 1.61}    & \textbf{91.35 $\pm$ 1.08}    & \textbf{66.20 $\pm$ 1.28}    \\
\bottomrule
\end{tabular}
\vspace{10pt}
\caption{\textbf{Object classification results} ($20$-class top-1 accuracy, \%) across the four heterogeneous tactile sensors. Overall averages available sensors, so optical-only and four-sensor averages are not directly comparable. Best result per sensor in \textbf{bold}; second-best \underline{underlined}.}
\label{tab:classification}
\end{table}
\vspace{-10pt}

From Table~\ref{tab:classification}, both MAE and HTT outperform \textit{Scratch} on every sensor, confirming that our recipe yields useful features. Excluding MAE, the strongest baselines are SPH on 9DTact and AnyTouch on GSMini; HTT improves by 12.61$\%$ and 14.99$\%$ respectively. T3 transfers poorly to 9DTact, suggesting that pretraining on a similar set of optical-based sensors does not generalize well to different optical-based sensors. 
Comparing HTT to MAE isolates the effect of cross-sensor alignment. Alignment improves accuracy on the optical-based sensors ($+4.8\%$ on 9DTact, $+2.8\%$ on GSMini), but underperforms on Xela ($-4.3\%$). We attribute this asymmetry to an information imbalance between the paired modalities: the rich optical-based signal benefits from the complementary force cues through alignment, whereas the coarser array-based signal has less room to be improved and can mildly drift toward its paired optical sensor's representation.

HTT pretraining yields useful features across all four heterogeneous tactile sensors (\textbf{Q1}). The alignment effect is mixed at the classification level --- consistent gains on the optical-based sensors but a small regression on Xela. We will revisit Q3 at Section~\ref{Sec: transfer}, where the effect is sharper.

\subsection{Transfer to Force and Slip Perception} \label{Sec: transfer}

\begin{table}[tbp]
\centering
\setlength{\tabcolsep}{4.3pt}
\small
\begin{tabular}{l ccccc ccccc}
\toprule
 & \multicolumn{5}{c}{\textbf{Force Estimation} (3D MAE, N\,$\downarrow$)} & \multicolumn{5}{c}{\textbf{Slip Detection} (Macro-F1, \%\,$\uparrow$)} \\
\cmidrule(lr){2-6}\cmidrule(lr){7-11}
\textbf{Method} & Xela & TAC-02 & 9DTact & GSMini & Overall & Xela & TAC-02 & 9DTact & GSMini & Overall\\
\midrule
Scratch     & 1.225             & 0.705             & 1.255             & 1.260             & 1.111             & 29.80             & 31.77             & 31.00             & 32.00             & 31.14         \\
T3          & n/a               & n/a               & 2.678             & 1.197             & 1.938             & n/a               & n/a               & 31.27             & 38.25             & 34.76          \\
SITR        & n/a               & n/a               & 1.085             & 1.373             & 1.229             & n/a               & n/a               & 41.77             & 36.36             & 39.07         \\
AnyTouch & n/a & n/a & 1.233 & 1.401 & 1.317 & n/a & n/a & 31.23 & 58.75 & 44.99 \\
SPH & 0.838 & 0.555 & 0.780 & 0.845 & 0.755 & \textbf{54.82} & 31.79 & 49.13 & \underline{71.86} & \underline{51.90} \\
MAE  & \underline{0.762} & \underline{0.516} & \textbf{0.574}    & \underline{0.803} & \underline{0.664} & \underline{54.46}    & \underline{33.45} & \underline{49.70} & 68.86 & 51.62\\
HTT  & \textbf{0.695}    & \textbf{0.508}    & \underline{0.606} & \textbf{0.736}    & \textbf{0.636}    & 54.21 & \textbf{45.45}    & \textbf{53.09}    & \textbf{72.65}    & \textbf{56.35}\\
\bottomrule
\end{tabular}

\vspace{10pt}
\caption{\textbf{Force-sensitive tasks results.} Force estimation reports the overall 3D force MAE (in N, lower is better); slip detection reports the macro-F1 (\%, higher is better) as the dataset is highly imbalanced. Full per-axis force results, per-class slip F1 and variance are reported in Appendix~\ref{App: results}}
\label{tab:perception_results}
\vspace{-20pt}
\end{table}

Both pretrained variants MAE and HTT  outperform Scratch, T3, SITR and AnyTouch for both tasks (Table~\ref{tab:perception_results}). T3 doubles the 9DTact force MAE relative to Scratch ($2.678$ vs $1.255$), SITR worsens force on GSMini ($1.373$ vs $1.260$), and they improve marginally on slip detection. This indicates their pretraining do not necessarily preserve features useful for force-sensitive tasks --- an issue our objective avoids because HTT pretraining is anchored to force-rich array data through cross-sensor pairing.
HTT also outperforms MAE on three of four sensors in both tasks, with the largest gain on slip detection ($+12.0$ macro-F1 on TAC-02). Slip detection relies on both force-sensitive cues and fine-grained spatial contact features: cross-modal alignment binds these complementary signals across paired optical and array sensors, yielding a representation that resists collapsing onto the dominant class under heavy imbalance.

Despite a significant distribution shift from the pretraining data, results show that the HTT representation transfers clearly to distinct tasks (\textbf{Q2}), where cross-sensor alignment improves the average results on these physics-grounded perception tasks (\textbf{Q3}).

\subsection{Shared-Space Analysis} \label{Sec: shared-space}

\begin{figure}[t]
\centering
\includegraphics[width=0.95\textwidth]{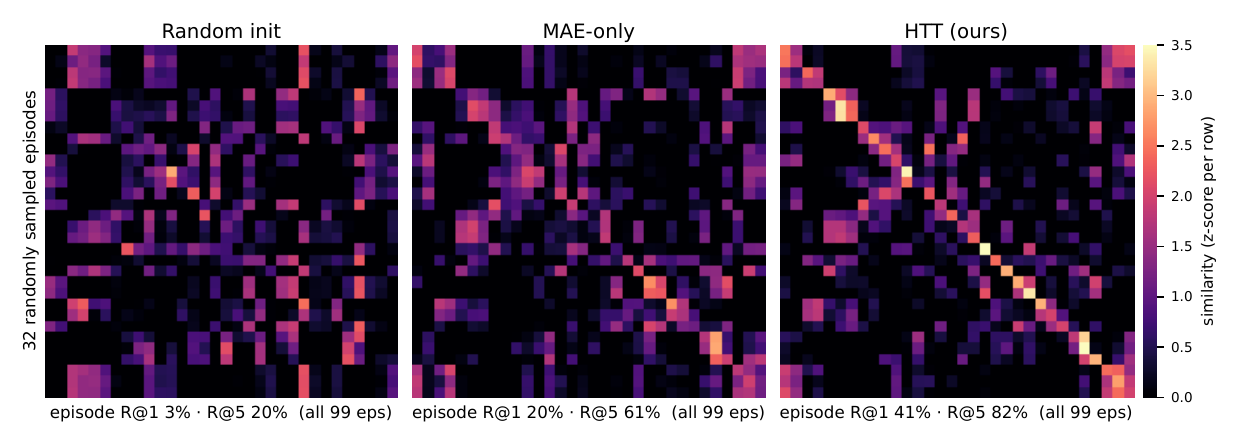}
\caption{\textbf{Cross-sensor retrieval between 9DTact and Xela after orthogonal Procrustes alignment.} Each panel shows similarities for 32 randomly sampled paired episodes. Paired counterparts lie on the diagonal; brighter diagonal entries indicate higher similarity for matching episodes. Colors are z-scored within each row for visualization. R@1/R@5 below each panel are evaluated over 99 episodes.}
\label{fig:cr-retrieval}
\end{figure}

We assess cross-sensor correspondence (\textbf{Q3}) by fitting an orthogonal Procrustes map to frozen trunk embeddings on half of the held-out paired contacts and testing on the disjoint half. Each 9DTact episode retrieves its Xela counterpart; R@$K$ is the fraction of queries with the counterpart among the top-$K$ matches. Across 99 episodes, R@1/R@5 increases from 3\%/20\% at random initialization to 20\%/61\% with MAE-only pretraining and 41\%/82\% with HTT (Figure~\ref{fig:cr-retrieval}), supporting improved cross-sensor correspondence under a fitted orthogonal map.

\subsection{HTT for Robot Manipulation}

\subsubsection{Real-World Experiments}

\textbf{Setup.} We deploy HTT on a Sharpa hand mounted on a Franka arm, with no external camera --- the policy must rely entirely on proprioception and finger-tip tactile sensing. We evaluate on two contact-rich tasks: \textit{toy screw}, where the fingers must repeatedly grip and rotate a screw until it is fully tightened; and \textit{grasp tofu}, where the hand must lift a piece of soft tofu off a plate without crushing it or letting it slip. We collect teleoperated demonstrations ($20$ for screw, $50$ for tofu) and train an ACT policy~\cite{Zhao-RSS-23}. The screw policy outputs hand-joint action chunks; the tofu policy additionally outputs the Franka $z$ position for the lift.

\textbf{Methods.} We compare five observation conditions: \textbf{qpos} uses only the $22$-D hand joint positions, with no tactile input; \textbf{wrench} concatenates qpos with the $6$-D force vector from each of the five finger-tip sensors; \textbf{HTT} concatenates qpos with HTT embeddings of the five finger-tip tactile images. T3 and SITR also adopts the same policy architecture, condition on qpos and their tactile embeddings. The Sharpa finger-tip tactile sensors are \emph{not} seen during HTT's pretraining; we apply the 9DTact encoder directly, without sensor-specific pretraining. Sharpa images are padded and resized to $224\times224$, background-subtracted using a no-contact reference, and normalized, with no sensor-specific calibration.

\begin{table}[htbp]
\centering
\small
\setlength{\tabcolsep}{6pt}
\begin{tabular}{lccccc}
\toprule
Task & qpos & +wrench & +T3 & +SITR & +HTT\\
\midrule
Toy screw & 2.5\% & 45\% & 0\% & 10\% & \textbf{90\%}\\
Grasp tofu & 5\% & 35\% & 5\% & 17.5\% & \textbf{55\%}\\
\bottomrule
\end{tabular}
\setlength{\abovecaptionskip}{7pt}
\setlength{\belowcaptionskip}{0pt}
\caption{\textbf{Real-world manipulation success rates}, evaluated over 40 rollouts per task and observation condition. Best result per task in \textbf{bold}.}
\label{tab:robot_main}
\end{table}

\begin{figure}[t]
    \centering
    \includegraphics[width=1.0\textwidth]{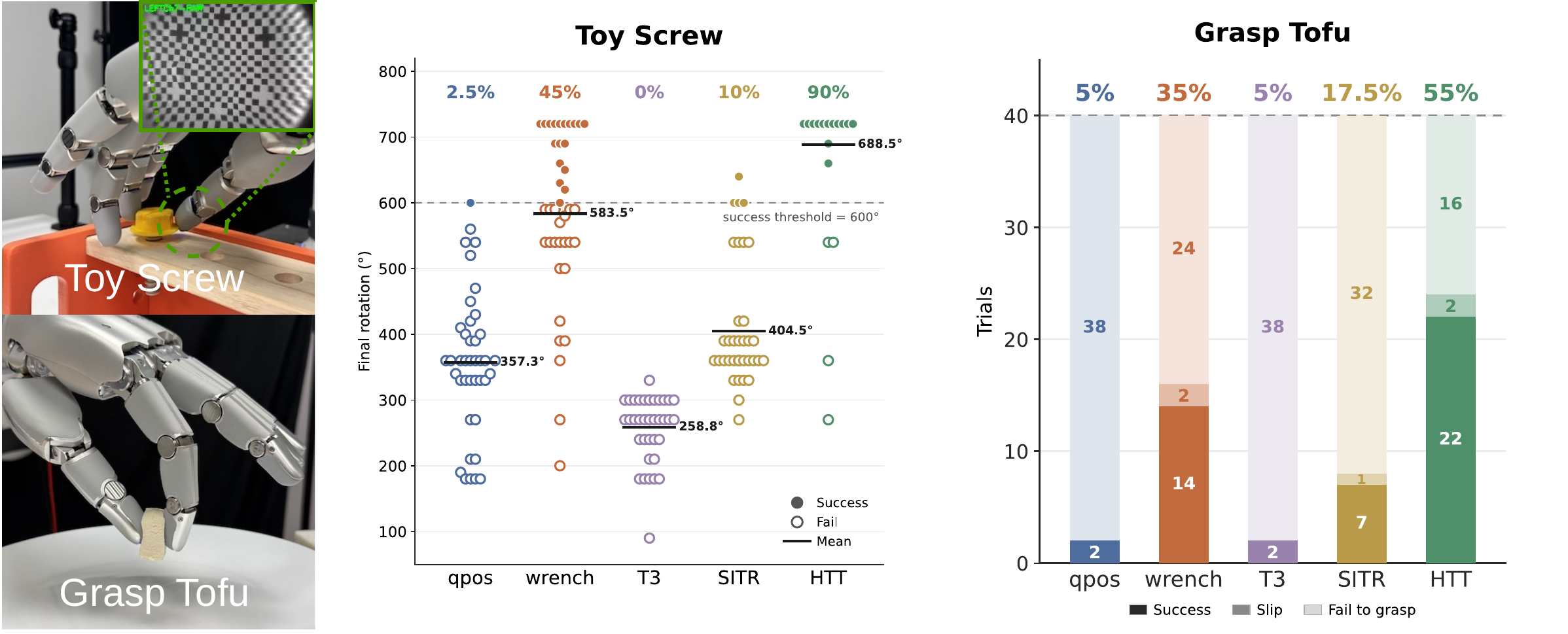}
    \vspace{-20pt}
    \caption{Real World Toy Screw and Grasp Tofu Experiments. Left: Task setup and tactile image from the Sharpa finger tip. Middle: Final rotation of screw; at least 600 degrees is considered success. Right: Grasp tofu completion status , showing success, slip, and failure to grasp. On both tasks, our HTT representations achieve the best results.}
    \label{Fig: robot}
    \vspace{-10pt}
\end{figure}

\textbf{Results.} On the \textit{toy screw} task, HTT achieves 90\% success (36/40), compared with 45\% for wrench and 0--10\% for the remaining baselines (Table~\ref{tab:robot_main}). Figure~\ref{Fig: robot} shows the final screw rotations and tofu grasp outcomes.
On the \textit{grasp tofu} task, both tactile conditions improve over qpos, with slip during the lift phase as the dominant failure mode (12/20 for wrench, 8/20 for HTT). Wrench crushes the tofu in 1/20 rollouts; HTT never crushes, but in one rollout fails to trigger the lift because the embedding does not register sufficient fingertip contact. Overall, HTT reduces the slip rate by one-third over the raw wrench, suggesting that its richer contact representation yields more reliable closed-loop control.

On two contact-rich manipulation tasks with unseen tactile sensors, HTT embeddings outperform baseline policies by large gains.
The results show that HTT provides richer tactile embeddings for policy learning than the sensor's raw force readout (\textbf{Q4}), and serves as a usable tactile backbone for new tactile sensors without sensor-specific pretraining (\textbf{Q5}).

\newpage
\subsubsection{ManiFeel Simulated Tasks}

\begin{wraptable}{r}{0.5\linewidth} 
\vspace{0pt}
\centering
\small                                 
\setlength{\tabcolsep}{4pt}            
\begin{tabular}{lcc}
\toprule
Method & Peg Insertion & Bulb Installation \\
\midrule
tacRGB & $0.21 \pm 0.02$          & $0.72 \pm 0.04$ \\
T3     & $0.23 \pm 0.02$          & $0.73 \pm 0.06$ \\
SITR   & $0.35 \pm 0.01$          & $\mathbf{0.77 \pm 0.04}$ \\
HTT(RGB)    & $\underline{0.44 \pm 0.04}$ & $\mathbf{0.77 \pm 0.02}$ \\
HTT(FF) & $\mathbf{0.48 \pm 0.12}$ & $0.76 \pm 0.02$ \\
\bottomrule
\end{tabular}
\caption{Task success rate averaged over three seeds, each with 50 rollouts.}
\label{tab:manifeel}
\end{wraptable}

In ManiFeel~\cite{luu2025manifeel}, we evaluate \textit{Peg Insertion} and \textit{Bulb Installation} with two tactile modalities: tactile RGB (TacRGB) and tactile force field (TacFF). We use GsMini encoder directly for tactile RGB data, and learn a TacFF encoder using TacFF data by initializing new encoder and decoder with pretrained shared chunk as backbone, and train with MAE loss.
From Table~\ref{tab:manifeel}, both HTT (RGB) and HTT(FF) outperforms baselines in \textit{peg insertion}. Performance converges near $0.77$ for all methods in \textit{bulb installation}, HTT is also among the best performing models. The results also confirm that HTT provides sensor-agnostic contact embeddings that boost policy learning (\textbf{Q4},\textbf{Q5}).

%% file: sections/6_conclusion.tex
\section{Conclusion}
We introduced the Heterogeneous Tactile Transformer (HTT), a self-supervised framework for learning shared representations across heterogeneous tactile sensors, together with the Heterogeneous Paired Tactile (HPT) dataset that supports such learning at scale. Across distinct perception and real-world manipulation tasks, HTT is shown to learn transferable representations that adapt to new tasks and unseen sensors. These results suggest that paired heterogeneous pretraining is a practical path toward general, sensor-agnostic tactile representation learning for perception and robot manipulation.

\section{Limitations and Future Work}
HTT leaves several questions open. First, our pretraining covers only optical-based and array-based tactile sensors; extending alignment to other sensing families (e.g., magnetic-based or fluid-based) is a natural next step. Second, our paired data is exclusively cross-family (optical $\leftrightarrow$ array), and the effect of pairing two sensors from the same family (e.g. two optical-based or two array-based) remains unexplored. Third, HPT pairs sensors in time and contact but not in geometric space: explicitly modeling which spatial patches of a optical sensor correspond to which taxels of an array sensor will be a future step for finer-grained cross-modal supervision.

%% file: sections/10_appendix.tex
\appendix

\section{Appendix}
\begin{figure}[ht]
    \centering
    \includegraphics[width=\textwidth]{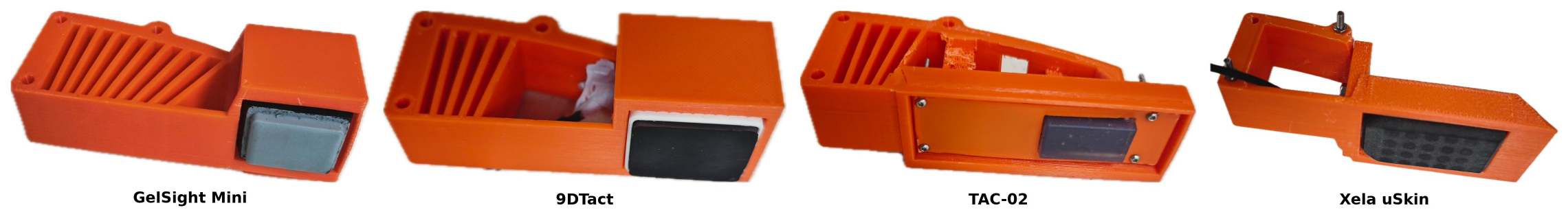}
    \caption{The four heterogeneous tactile sensors used in HPT pretraining, each mounted in the UMI gripper shell. Left to right: GelSight Mini and 9DTact (optical-based), TAC-02 and Xela uSkin (array-based).}
    \label{Fig: sensors}
\end{figure}
\subsection{Model and Training Details} \label{App: method}

\textbf{Architecture.} All modules share the same embedding dimension $D=192$ and use $3$ attention heads. The visual-tactile encoders and decoders each have depth $3$, taxel-sensor encoders have depth $2$. The shared transformer trunk has depth $9$. The cross-modal predictors each have depth $3$ and are implemented as stacks of cross-attention transformer blocks with learnable mask tokens. No CLS token is used; the trunk output retains the input token sequence.

\textbf{Patching.} optical-based inputs ($224\times 224 \times 3$, $2$ tubelets along time of size $2$) are split into non-overlapping spatial patches, yielding $196$ tokens per tubelet. Taxel inputs (20 frames or 40 frames) are split into non-overlapping temporal patches of length $4$, giving $5$ tokens (xela) or $10$ tokens (Tac-02) per sample. Both encoders are MAE-style: random masking is applied at the patch level.

\textbf{Masking ratios.} MAE reconstruction: $0.75$ for optical-based modalities and $0.60$ for taxel-based modalities. Cross-modal prediction: the predictor receives full source tokens and predicts the masked target tokens (per-modality target mask ratio of $0.80$ for taxel and $0.90$ for optical).

\textbf{Optimization.} AdamW, learning rate $3\times 10^{-4}$,  batch size $256$ paired samples per step. Linear warmup over the first $2{,}000$ steps from $3\times 10^{-6}$ followed by cosine decay back to $3\times 10^{-6}$ over the remaining steps. Gradients are clipped at $1.0$.

\textbf{Phased schedule.} A total of $50{,}000$ training steps. The weight of cross-modal alignment loss is initialized as $0$ and linearly increase along training steps for the first $20{,}000$ steps and fixed at $\alpha=0.1$ for the remaining steps.

\section{Baseline Details} \label{App: baselines}

\textbf{Scratch.} Sensor-specific models trained on each downstream task through supervised learning. We use the same per-modality encoder architectures as HTT but with more layers, train them jointly with the task heads, without any pretraining.

\textbf{T3.} A model pretrained on a 3M optical-based tactile dataset using sensor-specific encoders and a shared trunk. Since T3 only covers optical-based sensors, it is applied to GelSight Mini natively and to 9DTact in an out-of-distribution manner.

\textbf{SITR.} A model pretrained on a 1M frames of synthesized optical-based tactile data using a unified encoder for all tactile sensors. Inputs are background-subtracted before normalization to match SITR's training pipeline.

\textbf{MAE (ours).} Our model pretrained with unlabeled multi-modality data through per-modality MAE reconstruction only, without the cross-modal alignment loss.

\textbf{HTT (ours).} Our full model, jointly pretrained with per-modality MAE and cross-modal alignment loss using paired unlabeled tactile data collected by a UMI device.

\section{Detailed Experimental Results} \label{App: results}
AnyTouch and SPH results below report mean $\pm$ standard deviation across seeds 10, 20, and 30. Normal force uses $F_z$; shear averages errors over $F_x$ and $F_y$. SPH is the joint supervised HTT model trained without pretraining. Overall 3D-force MAE averages errors over $F_x$, $F_y$, and $F_z$. Slip class 0 (idle/no slip) corresponds to the static class below.

\textbf{Force estimation --- full per-axis results.} Table~\ref{tab:force_full} reports the shear, normal, and overall 3D force MAE where available; the main text (Table~\ref{tab:perception_results}) summarizes only the 3D MAE. Normal force is consistently the harder axis among methods with per-axis results.

\begin{table}[htbp]
\centering
\caption{\textbf{Force estimation full results} (test MAE in N, lower is better) for shear, normal, and overall 3D force. T3, SITR, and AnyTouch are optical-based methods (n/a for arrays). Best result per sensor and metric in \textbf{bold}; second-best \underline{underlined}.}
\label{tab:force_full}
\setlength{\tabcolsep}{4pt}
\small
\resizebox{\textwidth}{!}{%
\begin{tabular}{ll ccccccc}
\toprule
\textbf{Sensor} & \textbf{Force} & \textbf{Scratch} & \textbf{T3} & \textbf{SITR} & \textbf{AnyTouch} & \textbf{SPH} & \textbf{MAE (ours)} & \textbf{HTT (ours)} \\
\midrule
\multirow{3}{*}{Xela}
        & 3D     & 1.225 $\pm$ 0.218             & n/a                       & n/a                       & n/a & 0.838 $\pm$ 0.018 & \underline{0.762 $\pm$ 0.051} & \textbf{0.695 $\pm$ 0.033}    \\
        & Normal & 2.628 $\pm$ 0.642             & n/a                       & n/a                       & n/a & 1.779 $\pm$ 0.046 & 1.606 $\pm$ 0.147             & 1.376 $\pm$ 0.104             \\
        & Shear  & 0.524 $\pm$ 0.007             & n/a                       & n/a                       & n/a & 0.368 $\pm$ 0.005 & 0.340 $\pm$ 0.007             & 0.355 $\pm$ 0.005             \\
\midrule
\multirow{3}{*}{TAC-02}
        & 3D     & 0.705 $\pm$ 0.002             & n/a                       & n/a                       & n/a & 0.555 $\pm$ 0.007 & \underline{0.516 $\pm$ 0.022} & \textbf{0.508 $\pm$ 0.018}    \\
        & Normal & 2.125 $\pm$ 0.256             & n/a                       & n/a                       & n/a & 1.146 $\pm$ 0.020 & 1.040 $\pm$ 0.060             & 1.009 $\pm$ 0.049             \\
        & Shear  & 0.286 $\pm$ 0.002             & n/a                       & n/a                       & n/a & 0.260 $\pm$ 0.002 & 0.255 $\pm$ 0.004             & 0.257 $\pm$ 0.002             \\
\midrule
\multirow{3}{*}{9DTact}
        & 3D     & 1.255 $\pm$ 0.258             & 2.678 $\pm$ 0.468         & 1.085 $\pm$ 0.366         & 1.233 $\pm$ 0.273 & 0.780 $\pm$ 0.029 & \textbf{0.574 $\pm$ 0.005}    & \underline{0.606 $\pm$ 0.006} \\
        & Normal & 2.426 $\pm$ 0.741             & 6.265 $\pm$ 1.069         & 2.330 $\pm$ 0.866         & 2.679 $\pm$ 0.617 & 1.572 $\pm$ 0.084 & 1.042 $\pm$ 0.013             & 1.086 $\pm$ 0.009             \\
        & Shear  & 0.670 $\pm$ 0.033             & 0.884 $\pm$ 0.171         & 0.462 $\pm$ 0.117         & 0.511 $\pm$ 0.101 & 0.384 $\pm$ 0.002 & 0.339 $\pm$ 0.002             & 0.366 $\pm$ 0.007             \\
\midrule
\multirow{3}{*}{GSMini}
        & 3D     & 1.260 $\pm$ 0.021             & 1.197 $\pm$ 0.164         & 1.373 $\pm$ 0.148         & 1.401 $\pm$ 0.242 & 0.845 $\pm$ 0.009 & \underline{0.803 $\pm$ 0.092} & \textbf{0.736 $\pm$ 0.072}    \\
        & Normal & 2.503 $\pm$ 0.067             & 2.301 $\pm$ 0.353         & 2.357 $\pm$ 0.467         & 2.466 $\pm$ 0.544 & 1.427 $\pm$ 0.018 & 1.248 $\pm$ 0.180             & 1.045 $\pm$ 0.171             \\
        & Shear  & 0.639 $\pm$ 0.012             & 0.645 $\pm$ 0.105         & 0.881 $\pm$ 0.050         & 0.868 $\pm$ 0.093 & 0.554 $\pm$ 0.011 & 0.581 $\pm$ 0.048             & 0.582 $\pm$ 0.024             \\
\bottomrule
\end{tabular}%
}
\end{table}

\textbf{Slip detection --- full results.} Table~\ref{tab:slip_full} reports the overall macro-F1 together with the per-class F1 for the static, incipient, and slide classes. The from-scratch baseline collapses to the dominant \emph{slide} class on most sensors (near-zero F1 on the rare static and incipient classes), which inflates its slide-class F1 but yields a low macro-F1; HTT instead produces non-trivial F1 across all three classes.

\begin{table}[htbp]
\centering
\caption{\textbf{Slip detection full results} (test F1, \%, higher is better): overall macro-F1 and per-class F1 for the static, incipient, and slide classes. T3, SITR, and AnyTouch are optical-based methods (n/a for arrays). Best result per sensor and metric in \textbf{bold}; second-best \underline{underlined}.}
\label{tab:slip_full}
\setlength{\tabcolsep}{4pt}
\small
\resizebox{\textwidth}{!}{%
\begin{tabular}{ll ccccccc}
\toprule
\textbf{Sensor} & \textbf{Metric} & \textbf{Scratch} & \textbf{T3} & \textbf{SITR} & \textbf{AnyTouch} & \textbf{SPH} & \textbf{MAE (ours)} & \textbf{HTT (ours)} \\
\midrule
\multirow{4}{*}{Xela}
        & Macro-F1     & 29.80 $\pm$ 0.00          & n/a                        & n/a                        & n/a & 54.82 $\pm$ 1.18 & \textbf{54.46 $\pm$ 0.36}    & \underline{54.21 $\pm$ 0.79} \\
        & F1 static    & 0.0 $\pm$ 0.0             & n/a                        & n/a                        & n/a & 68.79 $\pm$ 3.51 & 64.01 $\pm$ 0.76             & 63.7 $\pm$ 0.0               \\
        & F1 incipient & 0.0 $\pm$ 0.0             & n/a                        & n/a                        & n/a & 2.38 $\pm$ 1.89 & 5.99 $\pm$ 0.15              & 6.2 $\pm$ 0.0                \\
        & F1 slide     & 89.4 $\pm$ 0.0            & n/a                        & n/a                        & n/a & 93.29 $\pm$ 1.60 & 93.39 $\pm$ 0.18             & 92.7 $\pm$ 0.0               \\
\midrule
\multirow{4}{*}{TAC-02}
        & Macro-F1     & 31.77 $\pm$ 0.00          & n/a                        & n/a                        & n/a & 31.79 $\pm$ 0.25 & \underline{33.45 $\pm$ 0.48} & \textbf{45.45 $\pm$ 7.27}    \\
        & F1 static    & 0.0 $\pm$ 0.0             & n/a                        & n/a                        & n/a & 0.22 $\pm$ 0.16 & 2.79 $\pm$ 1.30              & 41.3 $\pm$ 24.4              \\
        & F1 incipient & 0.0 $\pm$ 0.0             & n/a                        & n/a                        & n/a & 6.78 $\pm$ 0.19 & 7.63 $\pm$ 0.08              & 3.8 $\pm$ 3.8                \\
        & F1 slide     & 95.3 $\pm$ 0.0            & n/a                        & n/a                        & n/a & 88.37 $\pm$ 0.40 & 89.93 $\pm$ 0.09             & 91.3 $\pm$ 0.6               \\
\midrule
\multirow{4}{*}{9DTact}
        & Macro-F1     & 31.00 $\pm$ 0.00          & 31.27 $\pm$ 0.03           & 41.77 $\pm$ 6.69           & 31.23 $\pm$ 0.00 & 49.13 $\pm$ 3.45 & \underline{49.70 $\pm$ 0.42} & \textbf{53.09 $\pm$ 0.63}    \\
        & F1 static    & 0.0 $\pm$ 0.0             & 0.0 $\pm$ 0.0              & 14.7 $\pm$ 15.5            & 0.00 $\pm$ 0.00 & 34.69 $\pm$ 9.22 & 40.62 $\pm$ 1.07             & 39.1 $\pm$ 3.6               \\
        & F1 incipient & 0.0 $\pm$ 0.0             & 0.0 $\pm$ 0.0              & 19.2 $\pm$ 2.1             & 0.00 $\pm$ 0.00 & 22.86 $\pm$ 1.09 & 19.74 $\pm$ 1.24             & 30.6 $\pm$ 3.8               \\
        & F1 slide     & 94.0 $\pm$ 0.0            & 93.8 $\pm$ 0.1             & 91.5 $\pm$ 0.7             & 93.68 $\pm$ 0.00 & 89.83 $\pm$ 0.20 & 88.74 $\pm$ 0.42             & 89.6 $\pm$ 1.3               \\
\midrule
\multirow{4}{*}{GSMini}
        & Macro-F1     & 32.00 $\pm$ 0.00          & 38.25 $\pm$ 0.41           & 36.36 $\pm$ 0.12           & 58.75 $\pm$ 1.23 & 71.86 $\pm$ 0.24 & \underline{68.86 $\pm$ 1.93} & \textbf{72.65 $\pm$ 0.38}    \\
        & F1 static    & 0.0 $\pm$ 0.0             & 4.7 $\pm$ 0.5              & 0.0 $\pm$ 0.0              & 78.32 $\pm$ 5.92 & 81.82 $\pm$ 0.28 & 80.89 $\pm$ 1.79             & 83.02 $\pm$ 0.34             \\
        & F1 incipient & 0.0 $\pm$ 0.0             & 15.3 $\pm$ 1.8             & 13.1 $\pm$ 0.2             & 1.79 $\pm$ 2.53 & 37.76 $\pm$ 0.23 & 30.66 $\pm$ 4.19             & 38.62 $\pm$ 1.22             \\
        & F1 slide     & 95.0 $\pm$ 0.0            & 94.8 $\pm$ 1.1             & 96.0 $\pm$ 0.1             & 96.13 $\pm$ 0.31 & 96.01 $\pm$ 0.26 & 95.04 $\pm$ 0.72             & 96.31 $\pm$ 0.16             \\
\bottomrule
\end{tabular}%
}
\end{table}

\clearpage
\section{Dataset Related Details} \label{App: dataset}
\begin{figure}[ht]
    \centering
    \includegraphics[width=\textwidth]{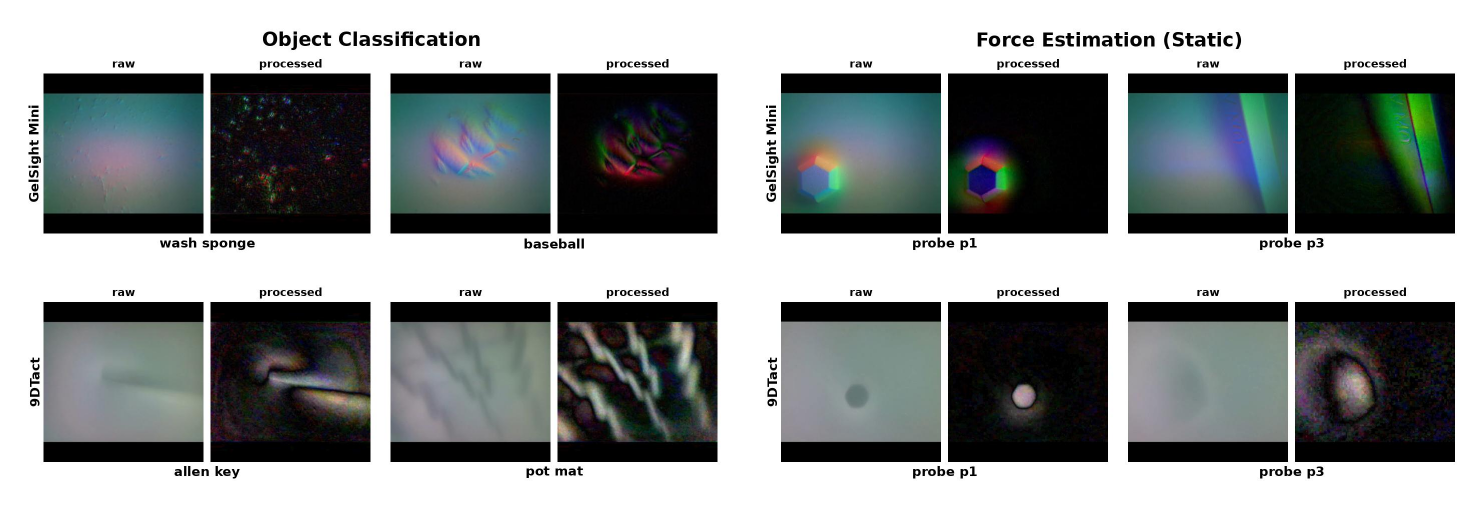}
    \caption{Example samples for 2 optical sensor encoders. \textbf{Left:} object classification frames (wash sponge, baseball on GelSight Mini; allen key, pot mat on 9DTact). \textbf{Right:} static force-estimation frames for two probe geometries ($p_1$, $p_3$) on each sensor. Each panel shows the raw camera frame and the background-subtracted processed frame; the processed view subtract with non-contact reference frames so amplifies the contact-induced deformation that the encoder operates on.}
    \label{Fig: samples}
\end{figure}

\subsection*{Object List for Classification}
\begin{itemize}[noitemsep, topsep=2pt]
    \item \textbf{Fruits/ Toys:} carrot, orange, lemon toy, cat bell, baseball, tennis ball.
    \item \textbf{Containers / lids:} glass cup, plastic cup, bottle lid.
    \item \textbf{Straps / clips:} metal watch strap, nylon watch strap.
    \item \textbf{Soft / textile:} baking glove, pot mat, table mat, washing sponge.
    \item \textbf{Stationery / tools:} colored notebook,  scratch paper, Allen key, sealing clip, screw
\end{itemize}
Sampled tactile images on classification and force estimation can be found in Figure~\ref{Fig: samples}.

\subsection*{Force Estimation Data}
\begin{itemize}
    \item Probe geometries: photos and physical dimensions of the four probes ($p_1$, $p_2$, $p_3$, $p_4$).
    \item Per-episode protocol: $4$-second duration, with the first $2$ frames used as a non-contact reference for calibration (background subtraction).
\end{itemize}

\subsection*{Slip Detection Labeling}
We derive per-frame slip labels post-hoc from the synchronized $6$-D force/torque signal. We first compute a friction-coefficient time series
\begin{equation}
\mu(t) = \frac{\sqrt{f_x(t)^2 + f_y(t)^2}}{|f_z(t)| + \epsilon}, \qquad \epsilon = 10^{-3},
\end{equation}
which captures the ratio of shear to normal force at each timestep. We then standardize $\mu(t)$ using a baseline (mean and standard deviation) estimated from the first $10\%$ of frames of each episode, during which the contact is assumed to be quasi-static. A two-sided Page CUSUM change-point detector~\cite{page1954continuous} is applied to the standardized signal to localize the onset of sliding: class~$2$ (gross slip) is triggered when the upper CUSUM statistic exceeds a threshold for several consecutive frames; class~$1$ (incipient slip) covers the few frames immediately preceding the trigger; all remaining frames are labeled class~$0$ (static).